\documentclass[10pt,twocolumn]{article} 
\usepackage{simpleConference}
\usepackage{times}
\usepackage{graphicx}
\usepackage{amssymb}
\usepackage{url,hyperref}
\usepackage{algorithm, algorithmic}
\usepackage{mathtools}
\usepackage[english]{babel}
\usepackage{amsthm}
 
\theoremstyle{definition}
\newtheorem{definition}{Definition}[section]

\begin{document}

\title{Multiple Manifold Clustering Using Curvature Constrained Path}

\author{Amir Babaeian \\
\\
University of California, San Diego, USA \\
\today
\\
\\
Email: ababaeian@ucsd.edu  \\
}

\maketitle
\thispagestyle{empty}

\begin{abstract}
The problem of multiple surface clustering is a challenging task, particularly when the surfaces intersect.
Available methods such as Isomap fail to capture the true shape of the  surface  near by the intersection and result in incorrect clustering.
The Isomap algorithm uses shortest path between points. The main draw back of the shortest path algorithm is due to the lack of curvature constrained where causes to have a path between points on different surfaces. In this paper we tackle this problem by imposing a curvature constraint to the shortest path algorithm used in Isomap. The algorithm chooses several landmark nodes at random and then checks whether there is a curvature constrained path between each landmark node and every other node in the neighborhood graph. We build a binary feature vector for each point where each entry represents the the connectivity of that point to a particular landmark. Then the binary feature vectors could be used as a input of conventional clustering algorithm such as hierarchical clustering. We apply our method to simulated and some real datasets and show, it performs comparably to the best methods such as K-manifold and spectral multi-manifold clustering.
\end{abstract}


\section{Introduction}
The application of unsupervised learning is considerably increased in different fields despite the current advancements in supervised learning (particularly deep learning \cite{Ali1, Ali2, Le1}).
We consider the problem of clustering points that are sampled in the vicinity of multiple surfaces embedded in Euclidean space, with a particular interest in the case where these intersect. The goal is multi-manifold clustering, which amounts to labeling each point according to the surface it comes from. 

This stylized problem may be relevant in a number of applications, such as the extraction of galaxy clusters \cite[]{MarSaa} and road tracking \cite{road-tracking} after some preprocessing.
In motion segmentation \cite{Ma07,1530127,vidal2006unified}, Human action recognition  \cite{yousefi2014comparative, yousefi2016slow,  yousefi2014development, yousefi2013biological, yousefi2018dual} and in face recognition \cite{Ho03,Basri03,Epstein95}, the underlying surfaces are usually assumed to be affine or, more generally, algebraic.

\begin{figure}[ht]
\begin{center}
\centerline{\includegraphics[width=4cm]{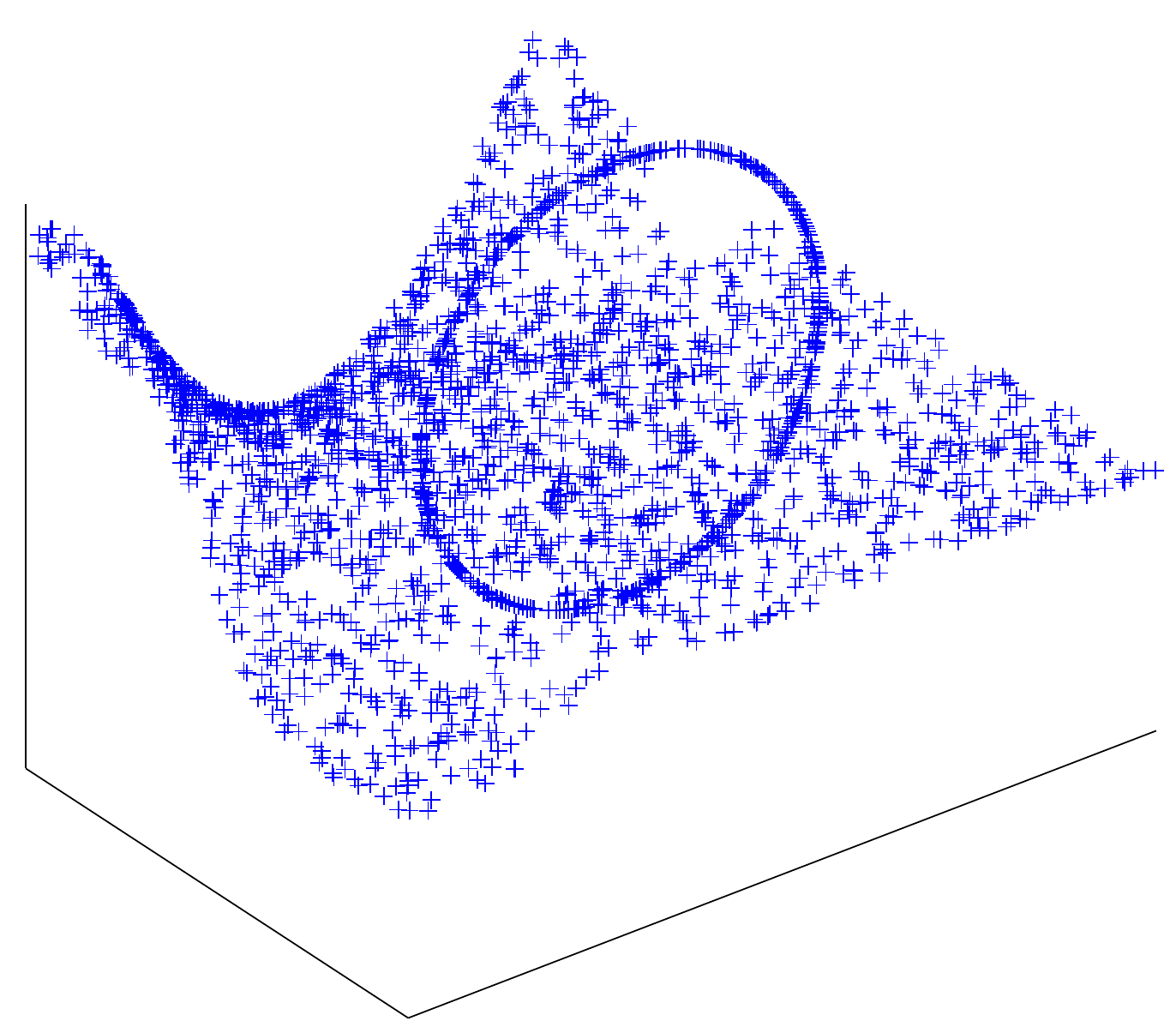}
\includegraphics[width=4cm]{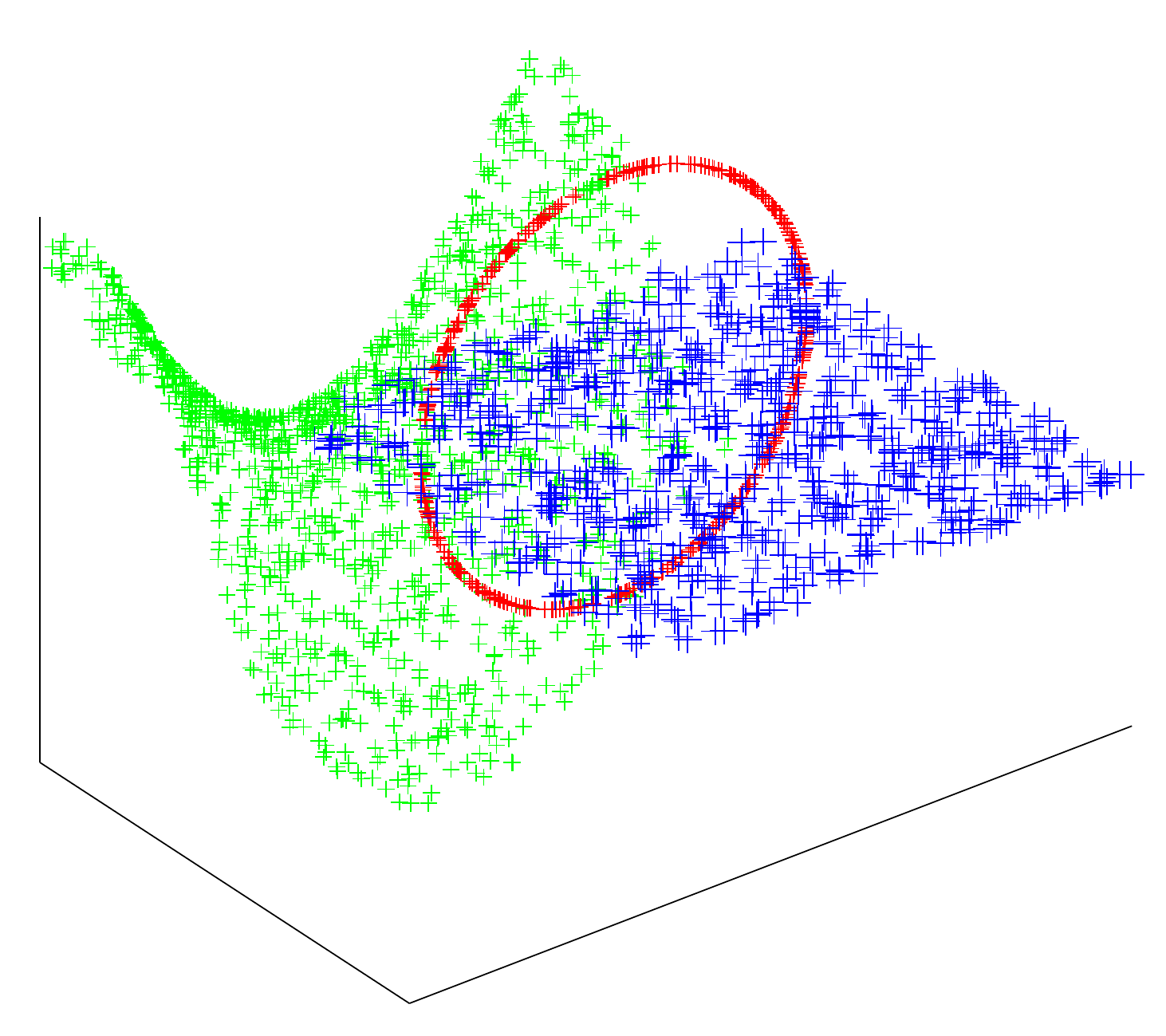}}
\caption{Simulated data illustrating the problem of multi-manifold clustering.  
Left: 3D data.  Right: output from our method.}
\label{fig:intro}
\end{center}
\vskip -0.2in
\end{figure}

Here we focus on a nonparametric setting where the main assumption is that the surfaces are smooth - see \ref{fig:intro} for an example.
This appears to be necessary to remove ambiguities in the problem of separating intersecting surfaces.

Several approaches have been proposed in this context.
Most methods are designed for the case where the surfaces do not intersect \cite{polito2001grouping,Ng02,NIPS2011_0065}, while others work only when the surfaces that intersect have different intrinsic dimension or density \cite{gionis,Haro06}.
The method of \cite{higher-order} is only able to separate intersecting curves.
Methods that purposefully aim at resolving intersections are fewer.
\cite{souvenir} implement some variant of K-means where the centers are surfaces.
\cite{energy} propose to minimize a (combinatorial) energy that includes local orientation information, using a tabu search.
The state-of-the-art method lies in methods based on local principal component analysis (PCA).
An early proposal was the elaborate multiscale spectral method of \cite{kushnir}, while the clustering routine of \cite{goldberg2009multi} --- developed in the context of semi-supervised learning --- inspired the works of \cite{wang2011spectral} and \cite{Gong2012}.

We propose a markedly different approach based on connecting points to landmarks via curvature-constrained paths.
It can be seen as a constrained variant of Isomap \cite{Tenenbaum00ISOmap}.
Isomap was specifically designed for dimensionality reduction in the single-manifold setting, and in particular, cannot handle intersections.
The curvature constraint on paths is there to prevent connecting points from one cluster to points from a different, intersecting cluster.
The resulting algorithm is implemented as a simple variation of Dijkstra's algorithm.
Our method is simpler than the previous proposals in the literature and performs comparably to the best methods, both on simulated and  real datasets.

The rest of the paper is organized as follows.
In \ref{sec:notion} we explain the notion of curvature constrained shortest-path and it's connection with the curvature constrained shortest-path.
In \ref{sec:alg} we present our algorithm for multi-manifold clustering and compare it with  three currently applied methods and give a theoretical guarantee for that.
In \ref{sec:num} we performed multiple numerical experiments on simulated and real data. Robustness of method to noise and choice of constraint is discussed as well.
In \ref{sec:diss} We discuss  and outline our future work  and development  of our algorithm.


\section{Constrained path} \label{sec:notion}
\subsection{Curvature constraint} \label{sec:notion}

Neighborhood graphs play a central role in manifold learning, exploiting the fact that smooth submanifolds are locally flat.  Recall that a neighborhood graph is a graph with vertices the sample points $x_1, \dots, x_N$.
We consider two types of neighborhood structure \cite{maier2009optimal}:
\begin{itemize}
\item {\em $\epsilon$-ball.}  $x_i$ and $x_j$ are connected if $\|x_i - x_j\| \le \epsilon$, where $\|\cdot\|$ denotes the Euclidean norm.
\item {\em $k$-nearest neighbor.}  $x_i$ and $x_j$ are connected if $x_j$ is among the $k$-nearest neighbors of $x_i$ (in the Euclidean metric), or vice-versa.
\end{itemize}

The central idea in this paper is the use of {\em constrained} shortest-path distances in a neighborhood graph.  The paths are constrained in order to control their smoothness.  The constrained shortest-path distances are used to estimate geodesic distances reliably, even when the surface self-intersects, thus allowing us to mimic Isomap.  
We use the fact that the constrained and unconstrained shortest-path distances are similar for points belonging to the same submanifold, while usually different for points belonging to different submanifolds.

For an ordered triplet of points $(x, y, z)$ in $\mathbb{R}^D$, we define the curvature as:

\[ \label{curv-def}
curv(x,y,z) = 
\begin{cases}
(R(x,y,z))^{-1}, & \text{if } \angle(x,y,z) < \frac\pi2, \\
\infty, & \text{otherwise},
\end{cases}
\]

 where $\angle$ stands for the angle and  $R(x,y,z)$ is the radius of the circle passing through $x,y,z$.

\begin{equation} \label{rad} R(x,y,z) = 
\frac{\sqrt{\splitfrac {\|x-y\|^2 + \|z-y\|^2}{+2 \|x-y\| \|z-y\| 
\cos \angle(x,y,z)}}}{\sin \angle(x,y,z)}.
\end{equation}

with $R(x,y,z) = \infty$ if $x,y,z$ are aligned.
 


\begin{definition} \label{curvature}
For a curvature $\kappa>0$, we say that a path $(x_{i_1}, \dots, x_{i_m})$ is $\kappa$-constrained if $curv(x_{i_{t-1}}, x_{i_t}, x_{i_{t+1}}) \le \kappa, \quad \forall t = 2, \dots, m-1$.
\end{definition}

%
%

To compute these constrained shortest-path distances we use a simple modification of Dijkstra's algorithm.  See \ref{alg:dijkstra} below.  When applied to a neighborhood graph with maximum degree $\Delta$, its computational complexity is $O(\Delta N \log N)$ per source point. 

\subsection{Angle constraint} \label{sec:notion}

For an ordered triplet of points $(x, y, z)$ in $\mathbb{R}^D$, define its angle as
\[ \label{angle-def}
 \splitfrac{\angle(x,y,z) = \angle(\Vec{xy}, \Vec{yz}) =\cos^{-1}\left(\frac{<y-x,z-y>}{\|y-x\| \|z-y\|}\right)}{\in [0, \pi]}
\]

We say that a sequence of points $(x_{i_{1}}, \dots, x_{i_{m}})$ is $\theta$-angle constrained if the angles between successive segments are all bounded by $\theta$, meaning
\[ \label{angle}
\angle(x_{i_{t-1}}, x_{i_{t}}, x_{i_{t+1}}) \le \theta, \quad \forall t = 2, \dots, m-1.
\]

\begin{figure}[htp]
\vskip 0.2in
\begin{center}
\centerline{\includegraphics[width=4.5cm]{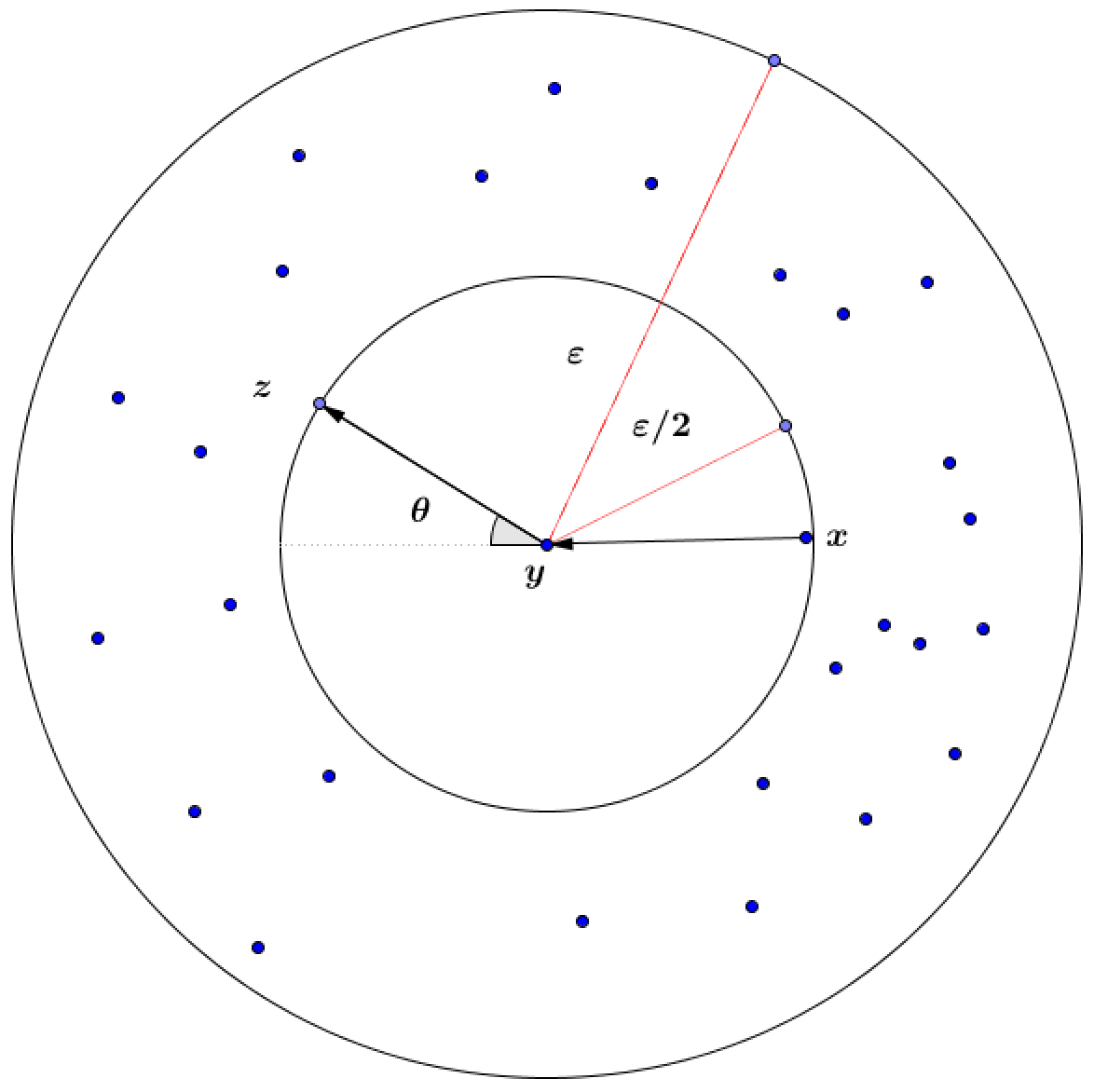}}
\caption{$x$ and $z$  lie in annulus neighborhood of point $y$.}
\label{fig:annulus}
\end{center}
\vskip -0.2in
\end{figure}

\ref{fig:annulus} shows three D-dimensional points  $x,y,z$ which form vertices of a triangle such that $x$ and $z$ belong to the annulus neighborhood of point $y$. Under above assumption the angle constraint  $\angle(x,y,z)<\theta$ where $\theta<\pi/2$ implies curvature constraint $curv(x,y,z)<\kappa$ where $\kappa=2sin(\theta )/\sqrt{\frac{\epsilon^{2}}{2}(1+cos(\theta))}$.

In  \ref{sec:intersect}  we analysis the correctness of our algorithm for simple case of two curves with intersection based on the angle constraint. 


\begin{algorithm}
   \caption{\quad Curvature Constrained Shortest Path Algorithm}
   \label{alg:dijkstra}
\begin{algorithmic}
   \STATE {\bfseries Input:} neighborhood graph $\boldsymbol{G}$ including weights of all the connected vertices, the landmark $x_\ell$ where $\ell=1,...,L$, angle or curvature constraint $\theta$.
   \FOR{$\ell=1$ {\bf to} $L$}
   
   \STATE $ss=\ell$, $t=\ell$, $m=0$.
    \STATE For each vertex $i$ of the graph($i=1, ..., N$): $distance[i]=Inf$, $cost[i]=Inf$, $parent[i]=Inf$, $temporary[i]=0$.
    \STATE At the beginning for each vertex $i$ of the graph, there is no path from that vertex to the landmark $x_\ell$, i.e., $path[\ell][i]=\left [ \ \right ]$.
   \FOR{$j=1$ {\bf to} $N$}
   \STATE Update $distance[i]$ for each $i\in neighbors(t)$ by the weight of edge between the vertex $i$ and $t$.
    \FOR{$i=1$ {\bf to} $N$}
    \IF{$distance[i]+m<cost[i]$}
    \IF{$ curv (i,t, ss)<\kappa$ or $t==ss$}
    \STATE Update $parent[i]=t$ and $cost[i]=distance[i]+m$.
   \ENDIF
   \ENDIF
   \ENDFOR
   \STATE Compute $temporary+cost$ vector then find the minimum element of these vector as well as the vertex $I$ with minimum element.
   \STATE Update $m$, i.e., $m=min(temporary+cost)$.
   \IF{$parent[i]\sim=Inf$}
   \STATE Update $path[\ell][I]$ by appending vertex $I$ to the end of $path[\ell][parent[I]]$.
   \ENDIF
   \STATE $temporary[I]=Inf$.
   \STATE $distance[i]=Inf$ for all $i=1, ..., N$.
   \STATE Update $t=I$ and choose $ss$ as the parent of t, i.e., $ss=parent[t]$.
   \STATE Update weights of edge from vertex $ss$ to $t$  and  from vertex $ss$ to $t$ by $Inf$ in order to avoid revisiting vertices. Our graph is  a directed graph, so it is possible to have edges in both directions between two vertices.
   \ENDFOR
\ENDFOR
\STATE {\bfseries Output:} Constrained shortest-Path from each vertex $i$ of the graph and each landmark $\ell$.

\end{algorithmic}
\end{algorithm}

\section{Multi-Manifold Clustering} \label{sec:alg}
\subsection{Existing methods}
The last decade saw a flurry of propositions aiming at clustering data points when the underlying clusters are not convex, and in particular, in the situation where the points are sampled near low-dimensional objects.
We gave a few references in the Introduction and now want to elaborate on three of them, \cite{souvenir}, \cite{spectral_applied} and \cite{wang2011spectral}, as we will use them as benchmarks in our experiments.
Our choice was dictated by performance, code availability and relevance to our particular setting.

The method of \cite{kushnir} renders impressive results but is hard to tune, having many parameters, while the method of \cite{Gong2012} is very similar to that of \cite{wang2011spectral} and the code was not publicly available at the moment of writing this paper.
The other methods for multi-manifold clustering that we know of were not designed to resolve intersections of clusters of possibly identical intrinsic dimensions and sampling densities.

We chose the subspace clustering method of \cite{spectral_applied} among a few others methods that perform well in this context.

\subsection{K-Manifolds}
\cite{souvenir} suggest an algorithm that mimics K-means, replacing centroid points with centroid submanifolds.
The method starts like Isomap by building a neighborhood graph and computing shortest path distances within the graph.  
After randomly initializing a $K$-by-$n$ weight matrix $W = (w_{ki})$, where $w_{ki}$ represent the probability that point $i$ belongs to the $k$th cluster, it alternates between an M-Step and an E-Step.
In the M-Step, for each $k$, the points are embedded in $\mathbb{R}^K$ using a weighted variant of multidimensional scaling using the weights $(w_{ki}: i =1,\dots,n)$.
In the E-Step, for each $k$ and $i$, the normal distance of point $x_i$ to the cluster $k$ is estimated as
\[\delta_{ki} = \frac{\sum_j w_{kj} (d(x_i, x_j) - d_k(x_i, x_j))}{\sum_j w_{kj}},\]
where $d(x_i, x_j)$ denotes the shortest path distance in the neighborhood graph and $d_k(x_i, x_j)$ denotes the Euclidean distance in the $k$th embedding, between points $x_i$ and $x_j$.
The weights are then updated as $w_{ki} \propto \exp(-d_{ki}^2/\sigma^2)$ such that $\sum_k w_{ki} = 1$ for all $i$, where $\sigma^2$ is chosen automatically.\\

\subsection{Spectral Curvature Clustering}

\cite{spectral_applied} proposed a spectral method for subspace clustering --- the setting where the underlying surfaces are affine.
We will compare our method to theirs when the surfaces are affine, and also when the surfaces are curved.  
The latter is done as a proof of concept, for it will be very clear that it cannot handle curved surfaces, like any other method for subspace clustering we know of.
The procedure assumes that all subspaces are of same dimension $d$, which is a parameter of the method.
For each $(d+2)$-tuple, $x_{i_1}, \dots, x_{i_{d+2}}$, it computes a notion of curvature $C_{i_1, \dots, i_{d+2}}$ which measure how well approximated this $(d+2)$-tuple is by an affine subspace of dimension $d$.
After reducing the tensor $\boldsymbol{C} = (C_{i_1, \dots, i_{d+2}}: i_t = 1, \dots, N)$ spectral graph partitioning \cite{Ng02} is applied.

\subsection{Spectral Multi-Manifold Clustering}

\cite{wang2011spectral} is a spectral method using a dissimilarity that factors in the Euclidean distance and the discrepancy between the local orientation of the data.
The surfaces are assumed to be of same dimension $d$ known to the user.
First, a mixture of probabilistic principal component analyzers \cite{tipping1999mixtures} are fitted to the data, approximating the point cloud by a union of $d$-planes.
This is used to estimate the tangent subspace at each data point.
The dissimilarity between two points is then an increasing function of their Euclidean distance and the principal angles between their respective affine subspaces.
These dissimilarities are fed to the spectral graph partitioning method of \cite{Ng02}.

\subsection{Our algorithm}
\label{sec:algorithm}

\renewcommand{\Vec}[1]{\overrightarrow{#1}}
We consider the following problem of surface clustering:

Given a sample $x_1, \dots, x_n \in \mathbb{R}^D$ sampled from $S_1 \cup \cdots \cup S_K$, where for each $k$, $S_k$ is a smooth, but possibly self-intersecting surface, label each point according to the surface it belongs to.

Our algorithm is quite distinct from all the other methods for multi-manifold clustering we are aware of, although it starts by building a $q$-nearest neighbor graph like many others.
The idea is very simple and amounts to clustering together points that are connected by an angle-constrained path in the neighborhood graph.

Take two surfaces $S_1$ and $S_2$ intersecting at a strictly positive angle.
Then for `most' pairs of data points $x_{i_1} \in S_1$ and $x_{i_2} \in S_2$, a path in the graph going from $x_{i_1}$ to $x_{i_2}$ has at least one large angle between two successive edges, on the order of the incidence angle between the surfaces; while for `most' pairs of data points $x_{i_1}, x_{i_2} \in S_1$, there is a path with all angles between successive edges relatively small.

To speedup the implementation, we select $M$ landmarks (with $M$ slightly larger than $K$) at random among the data points and only identify what data points are connected to what landmark via a $\kappa$-constrained path in the graph.  $M$ and $\kappa$ are parameters of the algorithm.

Let $\xi_{\ell i} = 1$ if point $i$ and landmark $\ell$ are connected that way, and $\xi_{\ell i} = 0$ if not.
We use $\boldsymbol{xi}_i := (\xi_{\ell i} : \ell =1, \dots, M)$ as feature vectors that we group together and cluster using hierarchical clustering with complete linkage.


%

\begin{algorithm}
   \caption{Path-Based Clustering (PBC)}
   \label{alg:clustering}
\begin{algorithmic}
   \STATE {\bfseries Input:} data $(x_i)$; parameters $q, K, M, \kappa$
   \STATE Build $q$-nearest neighbor graph
   \STATE Choose $M$ landmarks are random \\
   \FOR{$i=1$ {\bf to} $n$}
   \STATE For each landmark $\widehat{x}_\ell$, identify which points $x_i$ it is connected to via a $\kappa$-constrained path in the graph, and set $\xi_{\ell i} = 1$ if so, and $\xi_{\ell i} = 0$ otherwise.
   \ENDFOR
   \STATE Group and then apply hierarchical clustering to the feature vectors $\boldsymbol{xi}_1, \dots, \boldsymbol{xi}_n$ to find $K$ clusters, where $\boldsymbol{xi}_i := (\xi_{\ell i} : \ell =1, \dots, M)$.
\end{algorithmic}
\end{algorithm}


\subsubsection{Intersections}  \label{sec:intersect}
We are most interested in the case where the surfaces intersect.

Concretely, given $K$ compact, simply connected submanifolds $S_1, \dots, S_K \subset \mathbb{R}^D$ of maximum pointwise curvature bounded by $\kappa < \infty$, we consider the noisy mixture distribution
\[ \label{model0}
x = s + z, \quad s \sim \sum_{k=1}^K \pi_k \mu_{S_k} \quad z \sim \mu_{B(0,\tau)},
\]
where $\mu_S$ denotes the uniform distribution over set $S$.

\section{Numerical Expriments} \label{sec:num}

\subsection {Synthetic Data}
\begin{figure}[!ht]
\begin{center}
\centerline{
\includegraphics[width=0.45\textwidth]{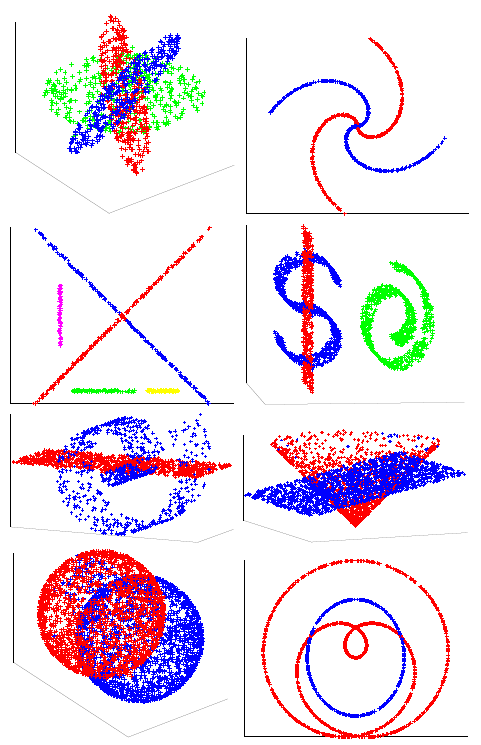}}
%
%
%

\caption{Result of our method on 8 synthetic datasets.}
\label{fig:Path-Synthetic}
\end{center}
\end{figure}

\begin{figure}[ht]
\vskip 0.2in
\begin{center}
\centerline{\includegraphics[width=3.5cm]{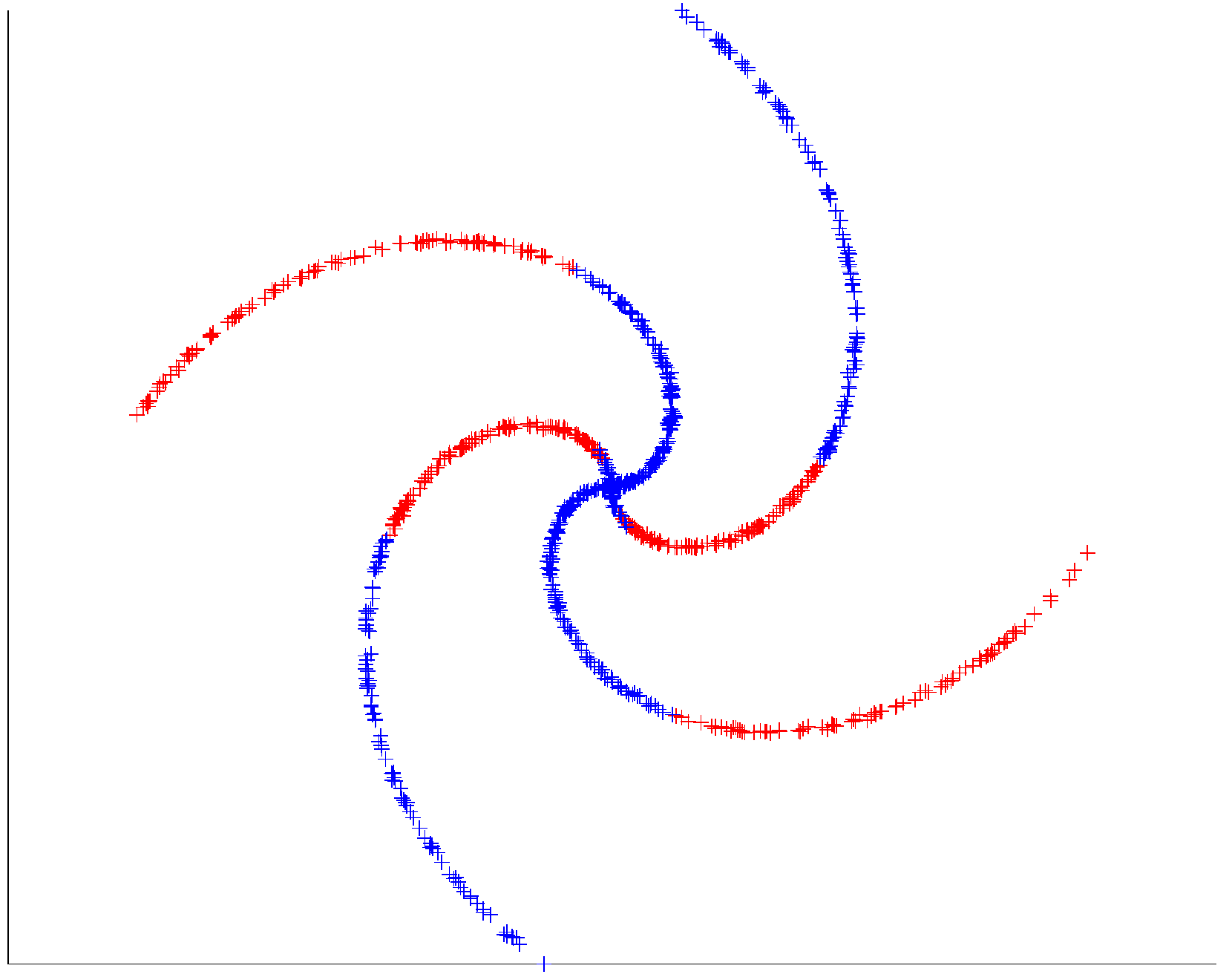}}
\caption{An example where SCC fails.}
\label{fig:scc-fail}
\end{center}
\vskip -0.2in
\end{figure}

\begin{figure}[ht]
\begin{center}
\centerline{\includegraphics[width=3.5cm]{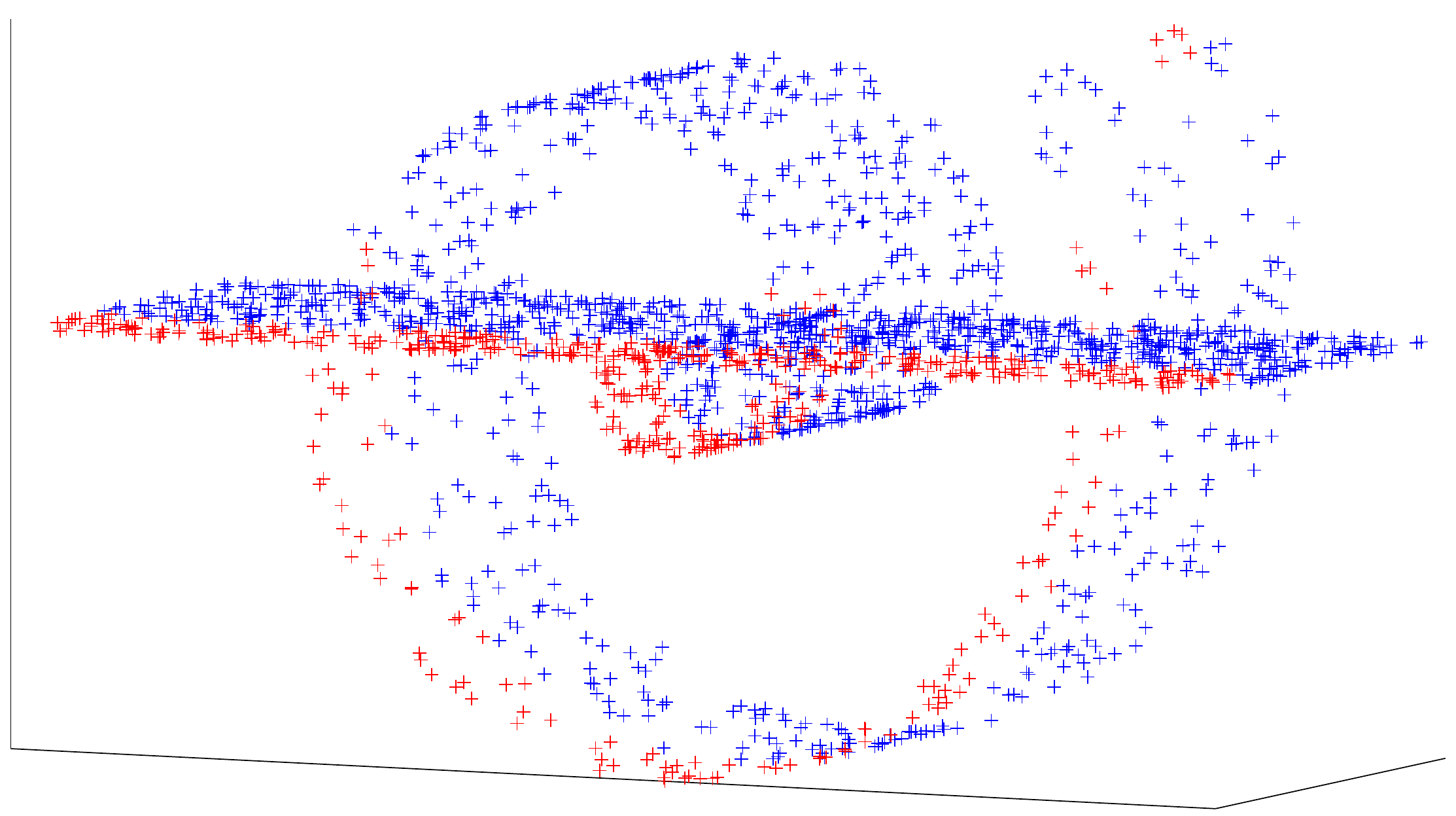}
\includegraphics[width=3.5cm]{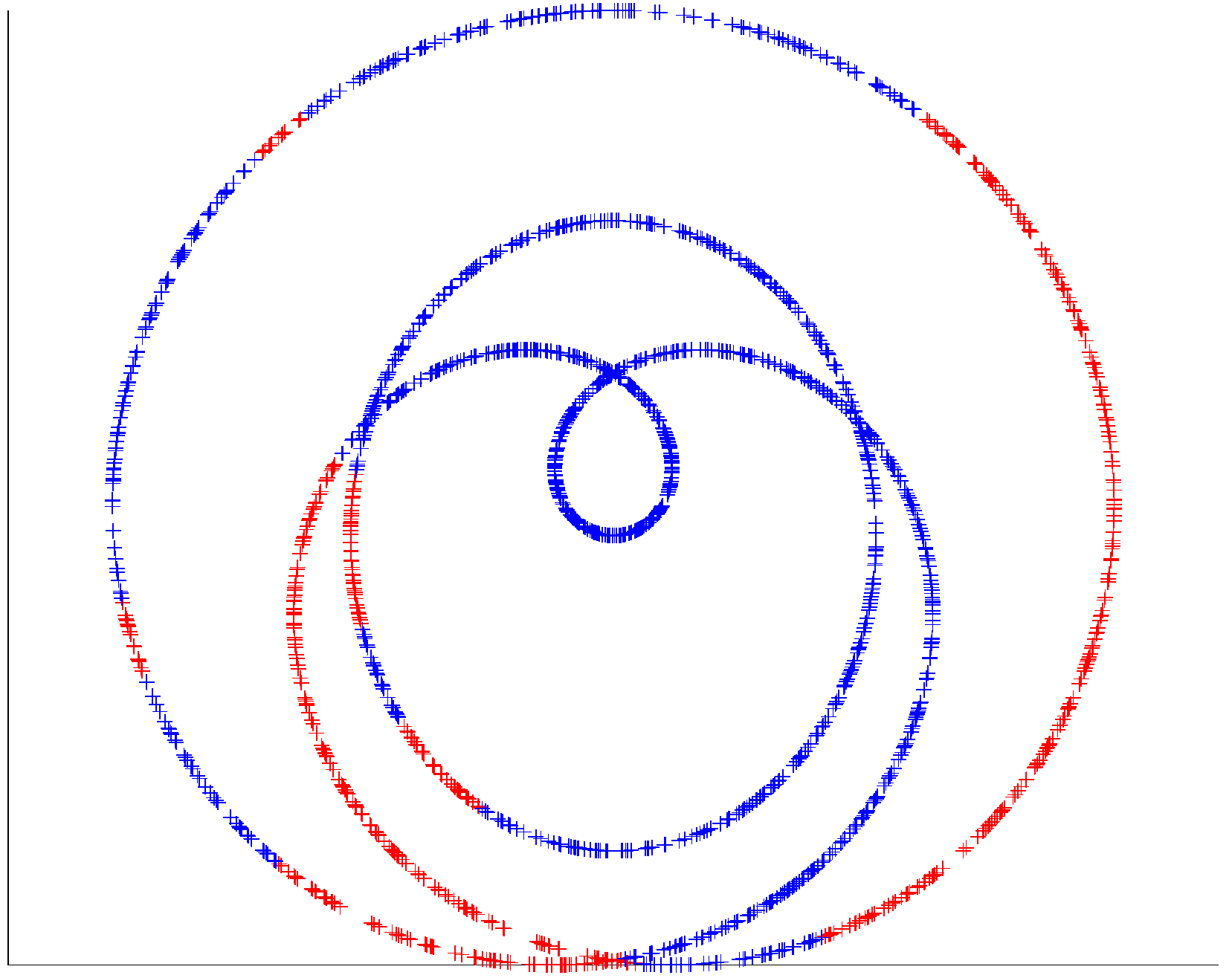}}
\caption{Examples where K-Manifolds fails.}
\label{fig:k-manifold-fail}
\end{center}
\vskip -0.2in
\end{figure}

\begin{figure}[h]
\begin{center}
\centerline{\includegraphics[width=3.5cm]{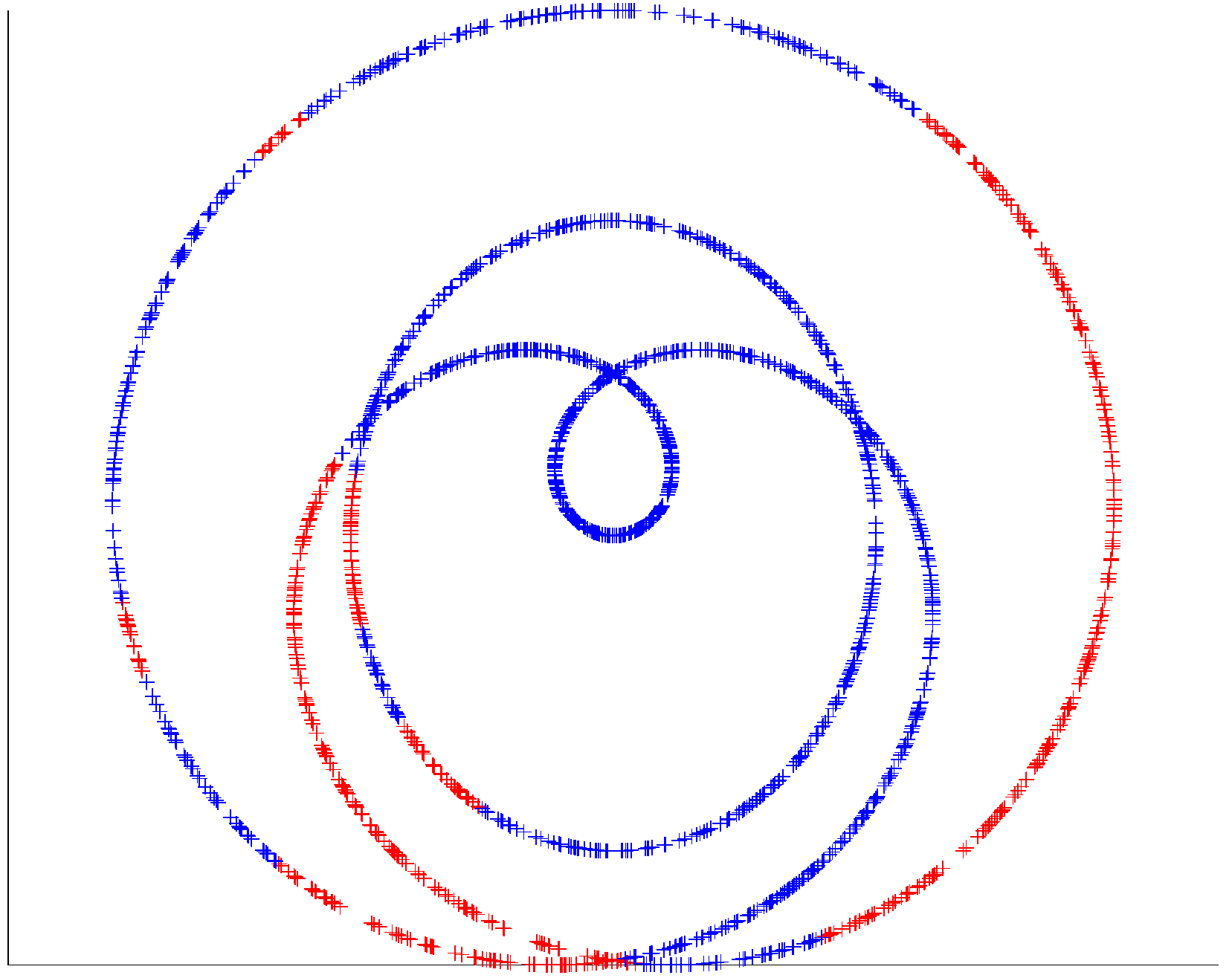}}
\caption{An example where SMMC fails.}
\label{fig:smmc-fail}
\end{center}
\vskip -0.2in
\end{figure}

The synthetic datasets we generated are similar to those appearing in the literature. \ref{fig:Path-Synthetic} shows the performance of our algorithm on eight synthetic data set.
The misclustering rates for our method, and the other three methods, are presented in Table~\ref{Compare}, where we see that our method achieves a performance at least comparable to the best of the other three methods on each dataset. We implemented path-based clustering using both curvature constraint and angle constraint with annuals graph. To compute the accuracy of clustering we remove a few ambiguous points close by intersection. 
Spectral Curvature Clustering (SCC) works well on linear manifolds (as expected) while it fails when there is curvature \ref{fig:scc-fail}. K-Manifolds fails in the more complicated examples \ref{fig:k-manifold-fail} . 
We found that this algorithm is very slow since it has to compute the shortest path between all the points, so that we could not apply it to some of the largest datasets. 
We mention that it assumes that clusters intersect, and otherwise does not work properly. 
Our method and Spectral Multi-Manifolds Clustering (SMMC) perform comparably on most datasets, but SMMC fails in the Rose Curve and Circle example  \ref{fig:smmc-fail} .
We note that K-Manifold, SCC and SMMC all require that all surfaces are of same dimension, which is a parameter of these methods, why our method does not need knowledge of the intrinsic dimensions of the surfaces and can operate even when these are different.
\vspace{-7mm}

\begin{table*}
\caption{Clustering accuracy on synthetic data.}
\label{Compare}
\begin{center}
\begin{small}
\begin{sc}
\begin{tabular}{lccccc}
\hline
Data set & K-Manifolds & SCC & SMMC & PBC-Angle-Annulus & PBC-Curvature \\
\hline
Three Planes                     & 97.1\% & 97.8\% & 99.5\%  & 99.6\% & 99.6\% \\
Two Spirals                       & 95.2\% & 54.8\% & 99.7\%  & 99.2\% & 99.1\% \\
Five Segments            & 59.1\% & 94.9\% & 99.6\%  & 98.1\%  & 98.0\%\\
Dollar-Sign, Plane and Roll       & 50.2\% &  -     & 99.6\%  & 99.7\% & 99.5 \% \\
Roll and Plane          & 56.5\% &  -     & 97.6\%  & 96.7\% & 96.9\% \\
Cone and Plane          & -      &  -     & 99.6\%  & 97.9\% & 98.1\% \\
Two Spheres                       & -      &  -     & 96.7\%  & 98.6\%  & 98.4\% \\
Rose Curve and Circle            & 62.9\% &  -     & 64.8\%  & 99.8\% & 99.7\% \\[.1in]
\hline
\end{tabular}
\end{sc}
\end{small}
\end{center}
\vskip -0.1in
\end{table*}
\vskip .5in

\subsection {Clustering of 2D Image Data}

In this section we apply our method on the COIL-20 dataset which includes 1440 gray-scale images of 20 objects. 
Each object contains 72 images taken by a camera at different angles. 
The original resolution of each image is $128\times128$. We first projected the dataset onto the top 10 principal components, then apply our path-based clustering algorithm. We tested our method on the three very similar objects 3, 6 and 19.
The algorithm is 99\% accurate (misclusters only 2 images out of 216) bringing a significant improvement over the state-of-the-art result of 70\% reported in \cite{wang2011spectral}. 
Lastly, we evaluated our method on the whole dataset obtaining an $83.6\%$ accuracy, improving on the 70.7\% accuracy reported in \cite{wang2011spectral}. 
(Here we used the top 20 principal components.) 
Since in this case we have 20 different classes, we increased the number of landmarks to 100 to make sure we sampled that at least a few landmarks from each class.

\begin{figure}[h]
\vskip 0.2in
\begin{center}
\centerline{\includegraphics[width=\columnwidth]{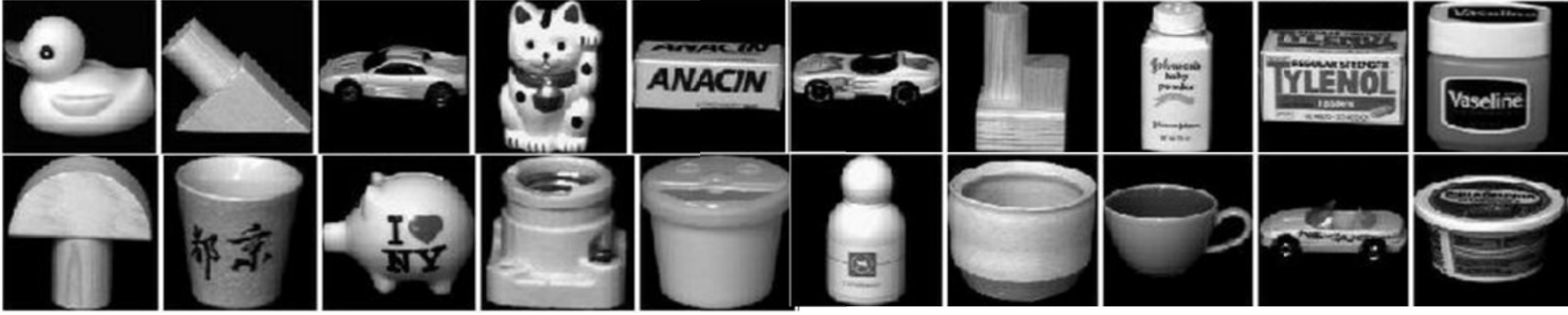}}
\caption{The 20 objects from the COIL-20 database.}
\label{fig:coil-data}
\end{center}
\vskip -0.2in
\end{figure}


\subsection {Clustering of Human Motion Sequences}
\begin{figure}[ht]
\vskip 0.2in
\begin{center}
\centerline{\includegraphics[width=\columnwidth]{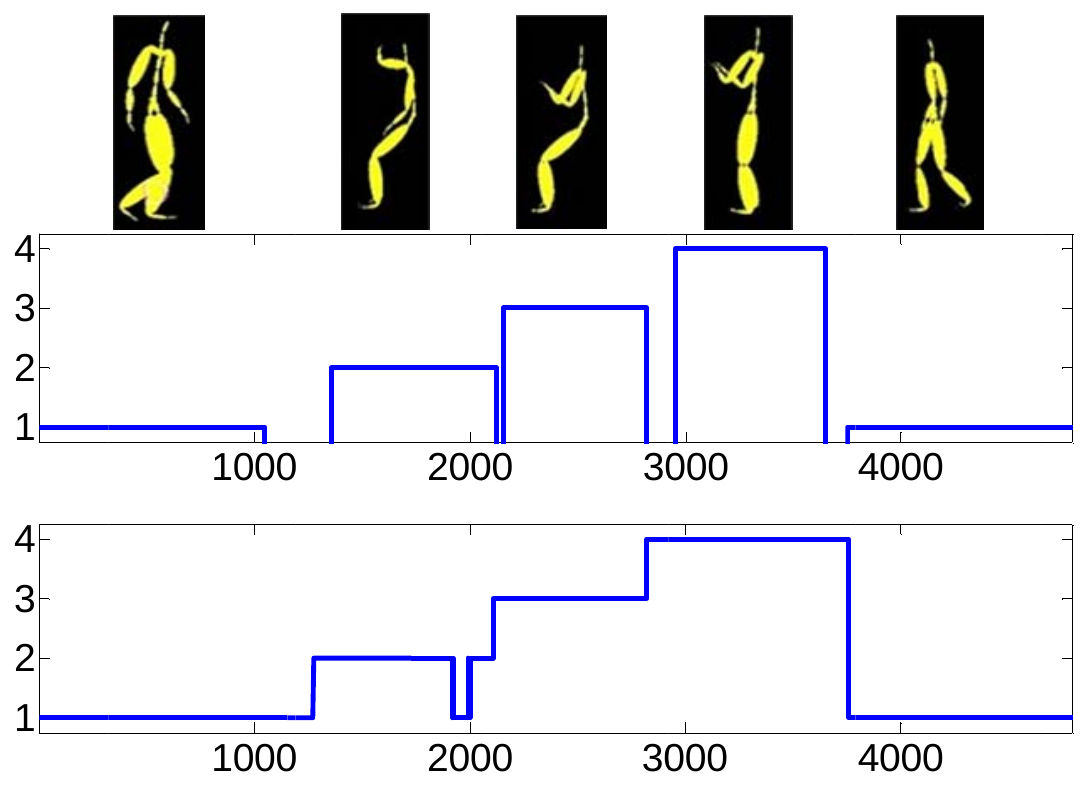}}
\caption{Result of human activity segmentation using Path-Based Clustering. 
There are 4 activities: walking (1), looking (2), sitting (3) and standing (4). 
Top: a sample of the sequence.  Middle: ground truth.  Bottom: output of our algorithm. }
\label{fig:human-activity}
\end{center}
\vskip -0.2in
\end{figure}

In computer vision clustering of human motion sequences into different class of activities performed by a subject is  referred to temporal segmentation. In this section we test our algorithm on a sequence of video frames including different activities performed by a subject.
We choose 4 mixed actions from subject 86, trial number 9 of the CMU MoCap dataset. 
The data consists in a temporal sequence of 62-dimensional representation of the human body via markers in $\mathbb{R}^3$. 
One motion sequence of 4794 frames and corresponding result of path-based multi-manifold clustering are given in \ref{fig:human-activity}. Four activities are labeled from 1 to 4. 
We do not label the frames where the subject switches from one action to another because of the uncertainty about the true activity.

\subsection {Segmentation of Video Sequences}

\begin{figure}[ht]
\vskip 0.2in
\begin{center}
\centerline{\includegraphics[width=\columnwidth]{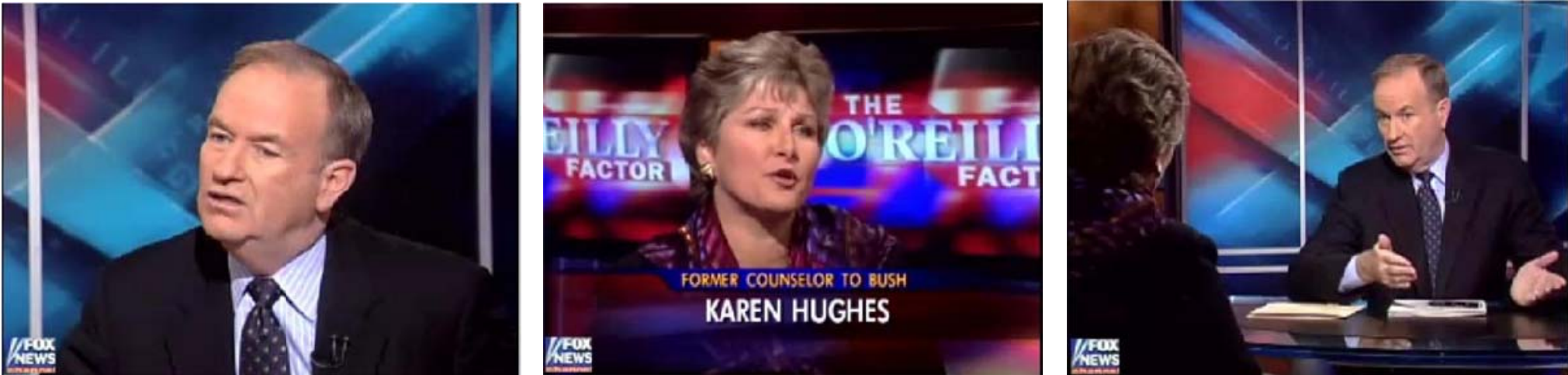}}
\caption{The first, 56th and last frame(135th) of Fox news video sequence.}
\label{fig:Fox news}
\end{center}
\vskip -0.2in
\end{figure}

In this section we consider the problem of partitioning a video sequence into different scenes. 
We consider the same video sequence used in \cite{spectral_applied,vidal2005generalized}. 
The video is an interview from Fox News containing 135 image frames of size $294\times413$; a sample is depicted in \ref{fig:Fox news}. 
Firstly we change each RGB image frame to the gray scale intensity image, then resize it to an $74\times104$ image. 
After concatenating all pixels of each image and putting into a vector of size $7696$, we construct a matrix of size $135\times7696$ where each row represents a frame of the original video sequence. 
Applying our algorithm on this matrix we get a perfect clustering (100\%). 
We repeated the experiment, this time projecting the data onto the top 10 principal components as done in \cite{spectral_applied,vidal2005generalized}, obtaining a matrix of size $135\times10$. 
We still get a 100\% accuracy, for an even wider range of parameters. 

\subsection{Robustness to Noise}
\begin{figure}[ht]
\vskip 0.2in
\begin{center}
\centerline{\includegraphics[width=\columnwidth]{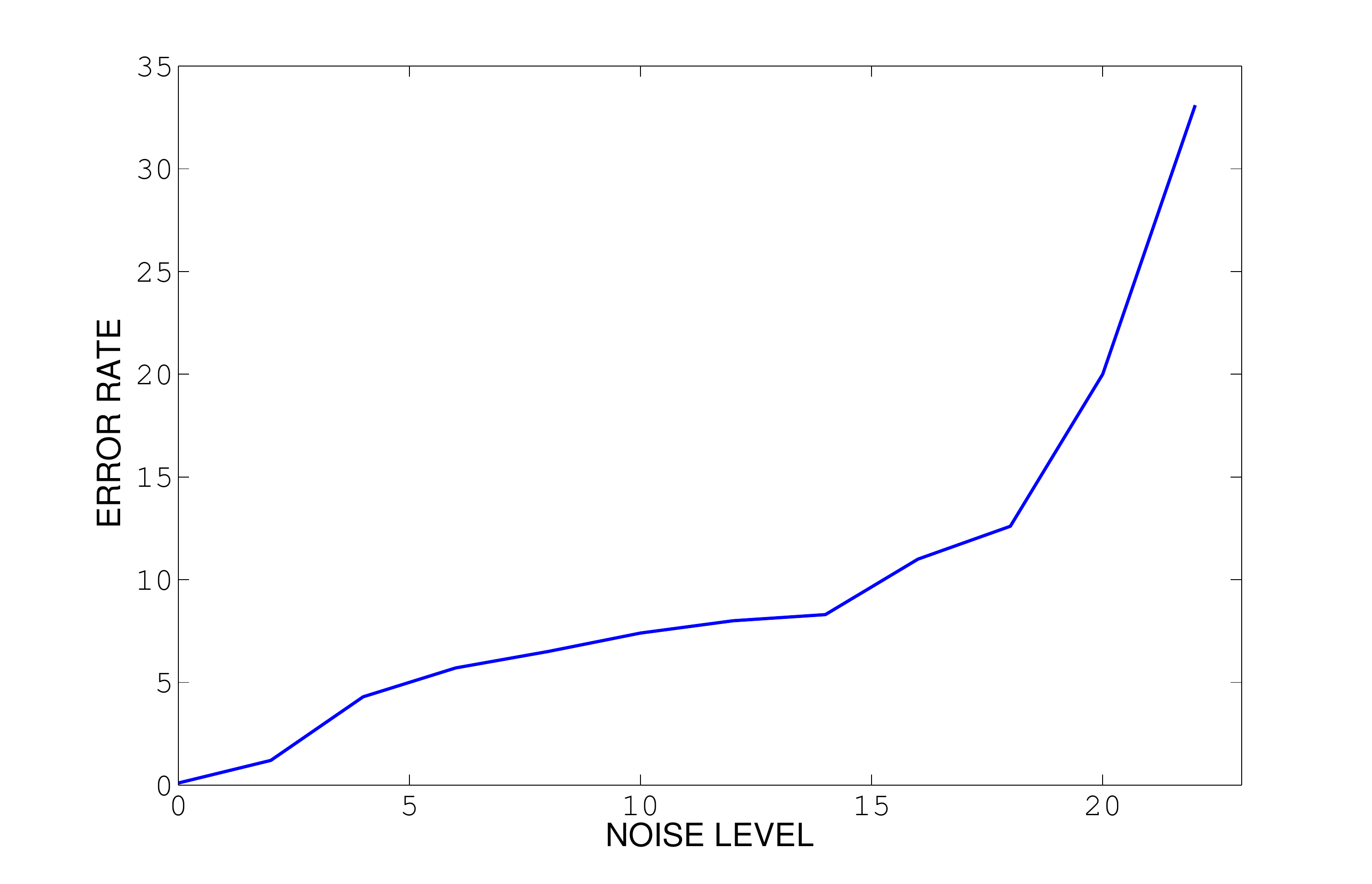}}
\caption{Effect of noise on performance of our algorithm on two intersecting curves shown in  \ref{fig:noisy} }
\label{fig:noise1}
\end{center}
\vskip -0.2in
\end{figure}

\begin{figure}[ht]
\centering
\begin{tabular}{cc}
\includegraphics[width=3cm]{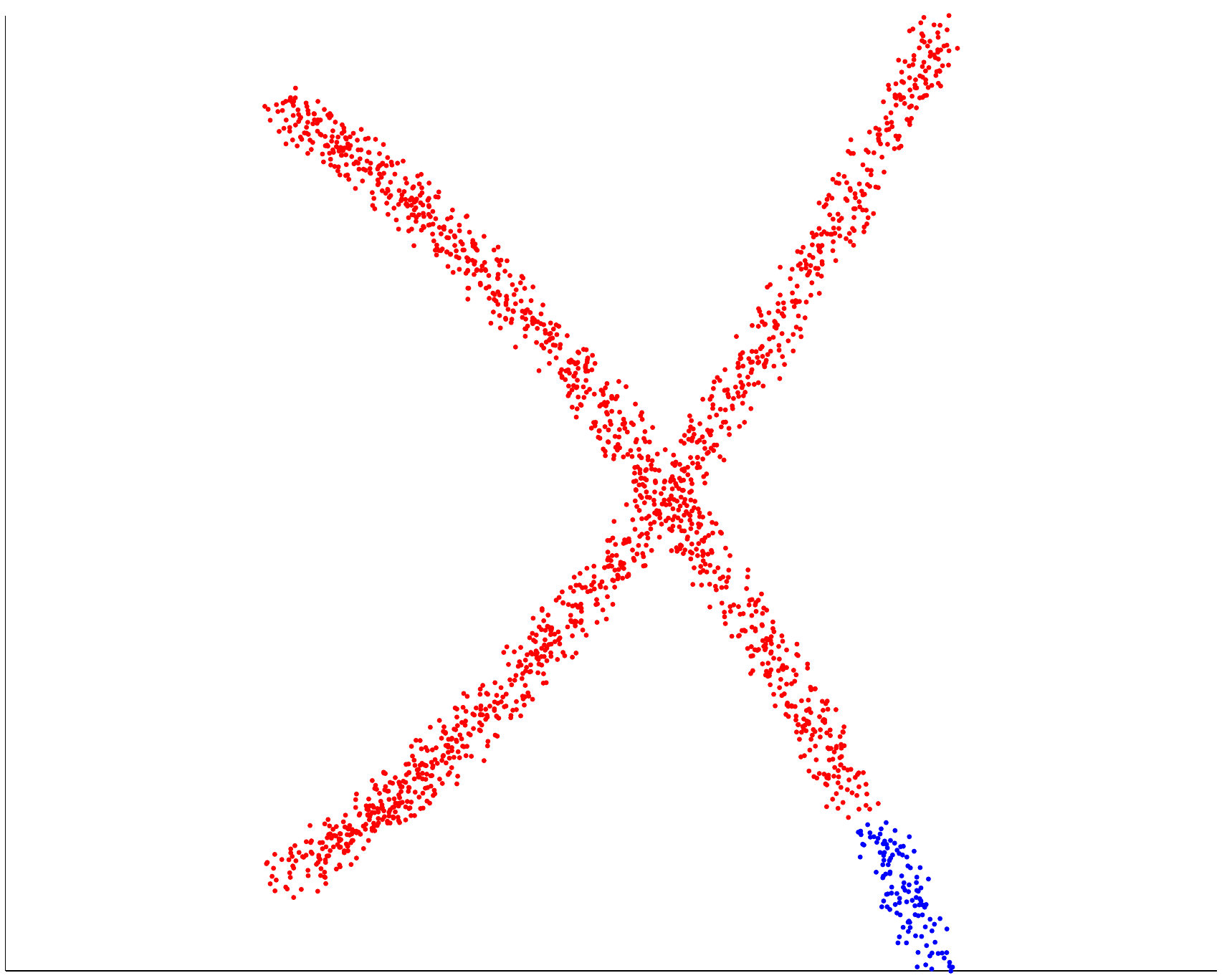} &
\includegraphics[width=3cm]{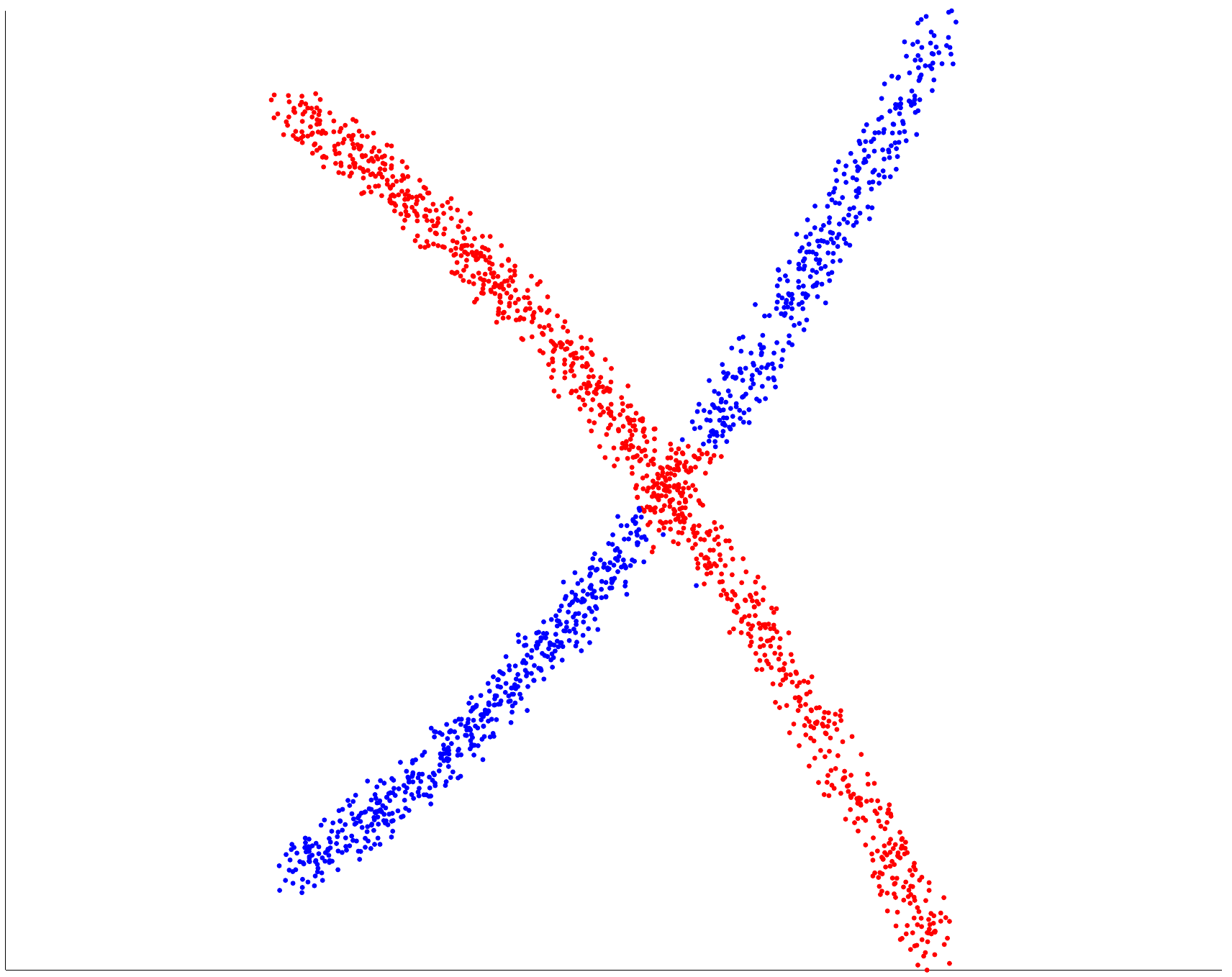} \\
Output of SMMC & Output of PBC
\end{tabular}
\caption{Example of noisy data.  
}
\label{fig:noisy}
\end{figure}

We note that all the other methods we know of for multi-manifold clustering do not perform well unless the noise level is quite small. 
As it appears in our method when we increase the amount of noise the possibility of connecting points from different surface with curvature constrained path increases. \ref{fig:noise1} shows the error rate for two intersecting curves with addition of standard uniform noise, as it can be illustrated, when we increase the  noise, notion of two different surface or manifold would be ambiguous where we can say all the points belong to one manifold. 
All other three methods  fail to capture the correct manifold with even small noise where our method still perform well with $20\%$ noise.
See \ref{fig:noisy} for an example with a substantial amount of noise, where SMMC fails while PBC succeeds.

\vspace{-7mm}

\subsection{The choice of constraint}
\begin{figure}[ht]
\centering
\begin{tabular}{cc}
\includegraphics[width=3.5cm]{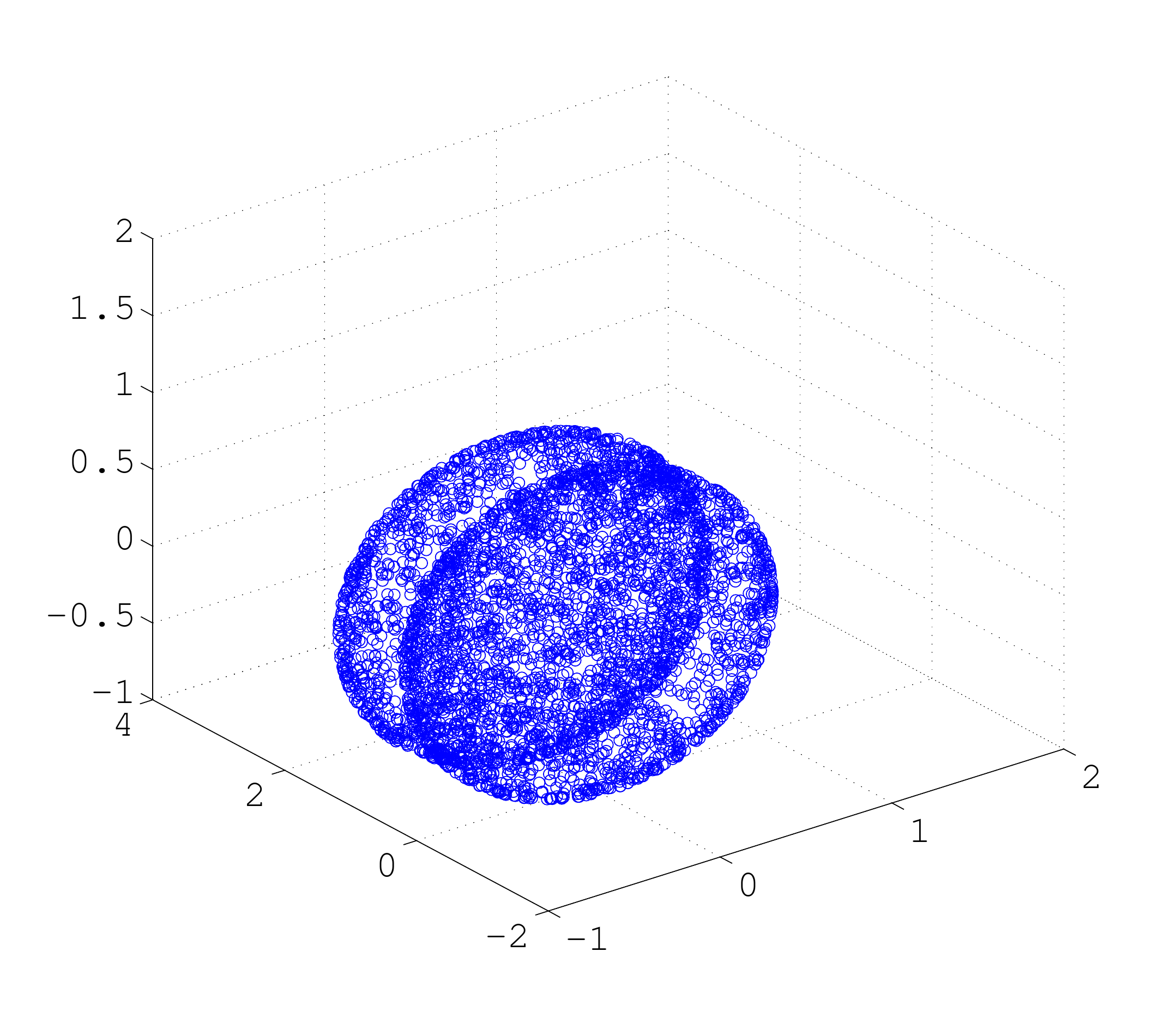} &
\includegraphics[width=3.5cm]{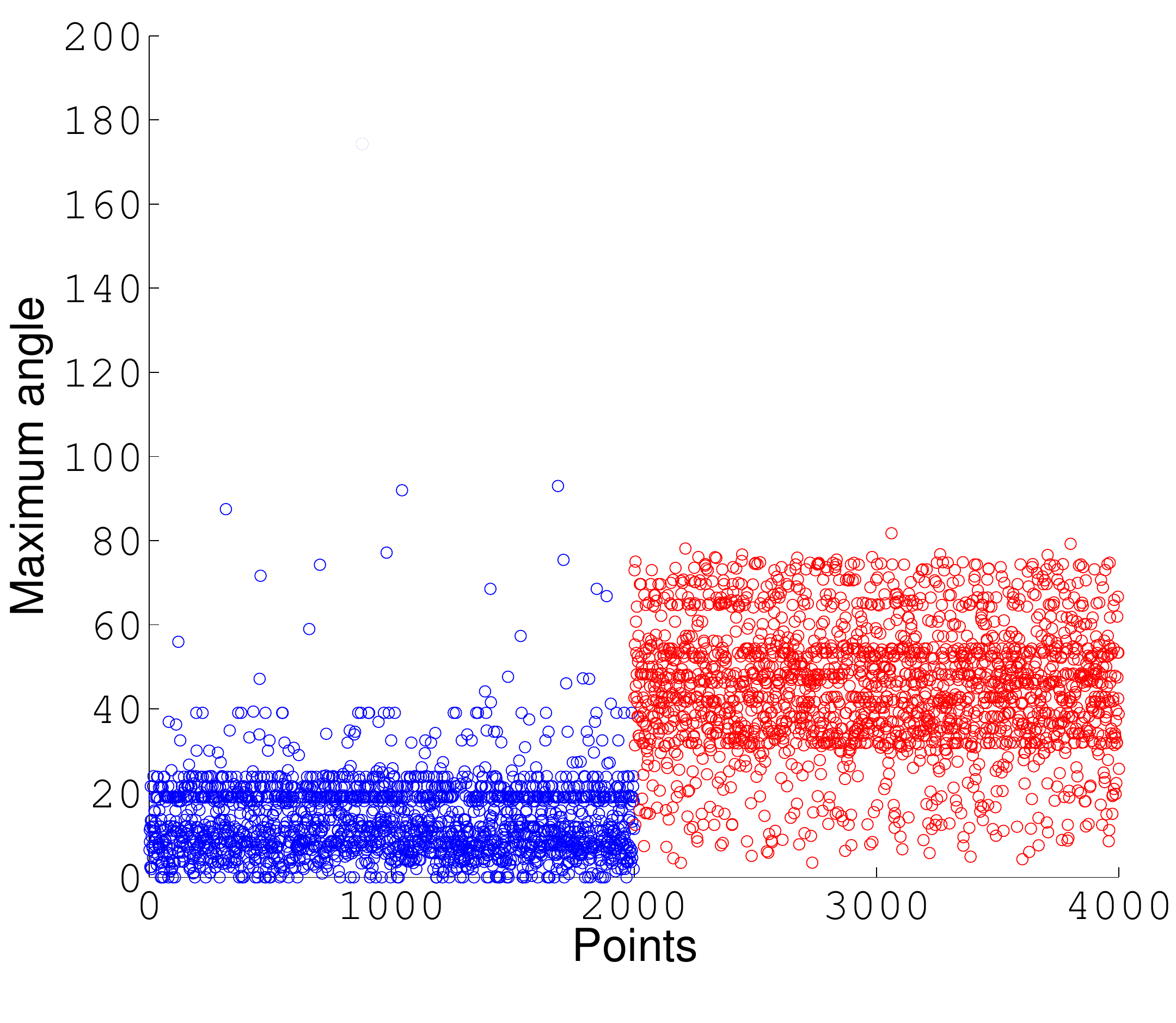} \\
\end{tabular}
\caption{left: two intersecting Sphere. Right: the distribution of maximum angle in unconstrained shortest-paths between points and  a given landmark.
}
\label{fig:angleanali}
\end{figure}

One of the main challenges of our algorithm is the choice of the angle constraint with annulus graph, since we deal with multiple manifolds with intersection. The  large angle constraint causes the points from different manifolds to  be connected using constrained shortest-path. This ultimately leads to multiple manifolds being clustered as one class.
We also considered implementing the small constraint, however, this constraint does not allow us to accurately capture the structure of the manifold. \ref{fig:angleanali} shows two intersecting spheres and the distribution of maximum angle in an unconstrained shortest-path between all the points and a given landmark. The distributions of maximum angle for the points within the same sphere as landmark belongs to (blue) is separable from the distributions of maximum angle for the points within the sphere that landmark does not belong to (red). This  illustration guides us to the idea that with the small amount of labeled points we are able to find the appropriate angle constraint. In another experiment we started with an angle constraint of $50^{\circ}$ and used $1\%$ of the points in each cluster as labeled data. We then compared the performance of our algorithm on the labeled data. In order to find the optimum angle constraint we  increased or decreased our angle constraint by a certain factor. We initially begin with dividing our angle constraint by a factor of $2$, until the error ceases to decrease. In the case that the error increases, we increase the angle constraint by a factor of $4/3$. In most cases we were able to find the optimal angle constraint within $5$ iterations. As it can be understood from \ref{fig:angleanali} the distribution of the maximum angle of the points within a class follows a flat distribution. By having a small number of labeled points we are able to capture the distribution of the maximum angle for the rest of the points in that class.

%
\vspace{-7mm}

\section{Discussion and Conclusion}  \label{sec:diss} 
We are currently experimenting with variants --- some based on other constraints --- that would lead to path-based clustering algorithms that perform at least as well in practice as \ref{alg:clustering}, and are consistent in the large-sample limit.

\bibliographystyle{abbrv}
\bibliography{egbib}

\begin{thebibliography}{10}

\bibitem{higher-order}
E.~Arias-Castro, G.~Chen, and G.~Lerman.
\newblock Spectral clustering based on local linear approximations.
\newblock {\em Electron. J. Statist.}, 5:1537--1587, 2011.

\bibitem{Basri03}
R.~Basri and D.~Jacobs.
\newblock Lambertian reflectance and linear subspaces.
\newblock {\em IEEE PAMI}, 25(2):218--233, 2003.

\bibitem{spectral_applied}
G.~Chen and G.~Lerman.
\newblock Spectral curvature clustering {(SCC)}.
\newblock {\em IJCV}, 81(3):317--330, 2009.

\bibitem{NIPS2011_0065}
E.~Elhamifar and R.~Vidal.
\newblock Sparse manifold clustering and embedding.
\newblock In J.~Shawe-Taylor, R.~Zemel, P.~Bartlett, F.~Pereira, and
  K.~Weinberger, editors, {\em Advances in Neural Information Processing
  Systems 24}, pages 55--63. 2011.

\bibitem{Epstein95}
R.~Epstein, P.~Hallinan, and A.~Yuille.
\newblock $5\pm 2$ eigenimages suffice: {A}n empirical investigation of
  low-dimensional lighting models.
\newblock In {\em IEEE Workshop on Physics-based Modeling in Computer Vision},
  pages 108--116, June 1995.

\bibitem{1530127}
Z.~Fu, W.~Hu, and T.~Tan.
\newblock Similarity based vehicle trajectory clustering and anomaly detection.
\newblock In {\em ICIP}, pages II--602--5, 2005.

\bibitem{road-tracking}
D.~Geman and B.~Jedynak.
\newblock An active testing model for tracking roads in satellite images.
\newblock {\em IEEE Trans. Pattern Anal. Mach. Intell.}, 18:1--14, 1996.

\bibitem{gionis}
A.~Gionis, A.~Hinneburg, S.~Papadimitriou, and P.~Tsaparas.
\newblock Dimension induced clustering.
\newblock In {\em KDD '05: Proceedings of the eleventh ACM SIGKDD international
  conference on Knowledge discovery in data mining}, pages 51--60, New York,
  NY, USA, 2005. ACM.

\bibitem{goldberg2009multi}
A.~Goldberg, X.~Zhu, A.~Singh, Z.~Xu, and R.~Nowak.
\newblock {Multi-manifold semi-supervised learning}.
\newblock In {\em AISTATS}, pages 169--176, 2009.

\bibitem{Gong2012}
D.~Gong, X.~Zhao, and G.~Medioni.
\newblock Robust multiple manifolds structure learning.
\newblock In {\em Proc.\ 29th Intl.\ Conf.\ on Machine Learning (ICML)}, 2012.

\bibitem{energy}
Q.~Guo, H.~Li, W.~Chen, I.-F. Shen, and J.~Parkkinen.
\newblock Manifold clustering via energy minimization.
\newblock In {\em ICMLA '07: Proceedings of the Sixth International Conference
  on Machine Learning and Applications}, pages 375--380, Washington, DC, USA,
  2007. IEEE Computer Society.

\bibitem{Haro06}
G.~Haro, G.~Randall, and G.~Sapiro.
\newblock {Stratification learning: Detecting mixed density and dimensionality
  in high dimensional point clouds}.
\newblock {\em Advances in Neural Information Processing Systems}, 19:553,
  2007.

\bibitem{Ho03}
J.~Ho, M.~Yang, J.~Lim, K.~Lee, and D.~Kriegman.
\newblock Clustering appearances of objects under varying illumination
  conditions.
\newblock In {\em CVPR}, pages 11--18, 2003.

\bibitem{kushnir}
D.~Kushnir, M.~Galun, and A.~Brandt.
\newblock Fast multiscale clustering and manifold identification.
\newblock {\em Pattern Recogn.}, 39(10):1876--1891, 2006.

\bibitem{Le1}
Y.~LeCun, Y.~Bengio, and G.~Hinton.
\newblock Deep learning.
\newblock {\em nature}, 521(7553):436, 2015.

\bibitem{Ma07}
Y.~Ma, A.~Y. Yang, H.~Derksen, and R.~Fossum.
\newblock Estimation of subspace arrangements with applications in modeling and
  segmenting mixed data.
\newblock {\em SIAM Review}, 50(3):413--458, 2008.

\bibitem{maier2009optimal}
M.~Maier, M.~Hein, and U.~von Luxburg.
\newblock Optimal construction of k-nearest-neighbor graphs for identifying
  noisy clusters.
\newblock {\em Theoretical Computer Science}, 410(19):1749--1764, 2009.

\bibitem{MarSaa}
V.~Mart\'inez and E.~Saar.
\newblock {\em Statistics of the Galaxy Distribution}.
\newblock CRC press, Boca Raton, 2002.

\bibitem{Ali1}
A.~Mortazi and U.~Bagci.
\newblock Automatically designing cnn architectures for medical image
  segmentation.
\newblock In {\em International Workshop on Machine Learning in Medical
  Imaging}, pages 98--106. Springer, 2018.

\bibitem{Ali2}
A.~Mortazi, R.~Karim, K.~Rhode, J.~Burt, and U.~Bagci.
\newblock Cardiacnet: Segmentation of left atrium and proximal pulmonary veins
  from mri using multi-view cnn.
\newblock In {\em International Conference on Medical Image Computing and
  Computer-Assisted Intervention}, pages 377--385. Springer, 2017.

\bibitem{Ng02}
A.~Ng, M.~Jordan, and Y.~Weiss.
\newblock On spectral clustering: Analysis and an algorithm.
\newblock {\em Advances in neural information processing systems}, 2:849--856,
  2002.

\bibitem{polito2001grouping}
M.~Polito and P.~Perona.
\newblock Grouping and dimensionality reduction by locally linear embedding.
\newblock {\em Advances in Neural Information Processing Systems},
  14:1255--1262, 2001.

\bibitem{souvenir}
R.~Souvenir and R.~Pless.
\newblock Manifold clustering.
\newblock In {\em Computer Vision, 2005. ICCV 2005. Tenth IEEE International
  Conference on}, volume~1, pages 648--653 Vol. 1, 2005.

\bibitem{Tenenbaum00ISOmap}
J.~B. Tenenbaum, V.~de~Silva, and J.~C. Langford.
\newblock A global geometric framework for nonlinear dimensionality reduction.
\newblock {\em Science}, 290(5500):2319--2323, 2000.

\bibitem{tipping1999mixtures}
M.~Tipping and C.~Bishop.
\newblock Mixtures of probabilistic principal component analysers.
\newblock {\em Neural Computation}, 11(2):443--482, 1999.

\bibitem{vidal2006unified}
R.~Vidal and Y.~Ma.
\newblock {A unified algebraic approach to 2-D and 3-D motion segmentation and
  estimation}.
\newblock {\em JMIV}, 25(3):403--421, 2006.

\bibitem{vidal2005generalized}
R.~Vidal, Y.~Ma, and S.~Sastry.
\newblock Generalized principal component analysis (gpca).
\newblock {\em Pattern Analysis and Machine Intelligence, IEEE Transactions
  on}, 27(12):1945--1959, 2005.

\bibitem{wang2011spectral}
Y.~Wang, Y.~Jiang, Y.~Wu, and Z.~Zhou.
\newblock Spectral clustering on multiple manifolds.
\newblock {\em Neural Networks, IEEE Transactions on}, 22(7):1149--1161, 2011.

\bibitem{yousefi2014comparative}
B.~Yousefi and C.~K. Loo.
\newblock Comparative study on interaction of form and motion processing
  streams by applying two different classifiers in mechanism for recognition of
  biological movement.
\newblock {\em The Scientific World Journal}, 2014, 2014.

\bibitem{yousefi2014development}
B.~Yousefi and C.~K. Loo.
\newblock Development of biological movement recognition by interaction between
  active basis model and fuzzy optical flow division.
\newblock {\em The Scientific World Journal}, 2014, 2014.

\bibitem{yousefi2016slow}
B.~Yousefi and C.~K. Loo.
\newblock Slow feature action prototypes effect assessment in mechanism for
  recognition of biological movement ventral stream.
\newblock {\em International Journal of Bio-Inspired Computation},
  8(6):410--424, 2016.

\bibitem{yousefi2018dual}
B.~Yousefi and C.~K. Loo.
\newblock A dual fast and slow feature interaction in biologically inspired
  visual recognition of human action.
\newblock {\em Applied Soft Computing}, 62:57--72, 2018.

\bibitem{yousefi2013biological}
B.~Yousefi, C.~K. Loo, and A.~Memariani.
\newblock Biological inspired human action recognition.
\newblock In {\em Robotic Intelligence In Informationally Structured Space
  (RiiSS), 2013 IEEE Workshop on}, pages 58--65. IEEE, 2013.

\end{thebibliography}
\end{document}